\documentclass{egpubl}
 
\JournalSubmission    
\NoStandardLicense

\usepackage[T1]{fontenc}
\usepackage{dfadobe}  

\usepackage{cite}  
\BibtexOrBiblatex
\electronicVersion
\PrintedOrElectronic
\ifpdf \usepackage[pdftex]{graphicx} \pdfcompresslevel=9
\else \usepackage[dvips]{graphicx} \fi

\usepackage{egweblnk} 
\usepackage{lineno}
\usepackage{graphicx}
\usepackage{amsmath}
\usepackage{amssymb}
\usepackage{booktabs}

\title[Infinite 3D Landmarks]%
      {Infinite 3D Landmarks: Improving Continuous 2D Facial Landmark Detection}

\author[P. Chandran \& G. Zoss \& P. Gotardo \& D. Bradley]
{\parbox{\textwidth}{\centering P. Chandran\orcid{0000-0001-6821-5815} \hspace{5mm} G. Zoss\orcid{0000-0002-0022-8203} \hspace{5mm} P. Gotardo\orcid{0000-0001-8217-5848}\thanks{Now at Google} \hspace{5mm} D. Bradley\orcid{0000-0002-2055-9325} 
        }
\\
{\parbox{\textwidth}{\centering 
\begin{tabular}{cc}
     DisneyResearch|Studios
\end{tabular}
       }}}

\usepackage{color}

\newcommand{\figref}[1]{Fig.~\ref{#1}}
\newcommand{\tabref}[1]{Table~\ref{#1}}
\newcommand{\eqnref}[1]{Eq.~\ref{#1}}
\newcommand{\secref}[1]{Section~\ref{#1}}

\newcommand{\etal}{{et al.}}
\begin{document}

\teaser{
   \vspace{-10mm}
   \includegraphics[width=\linewidth]{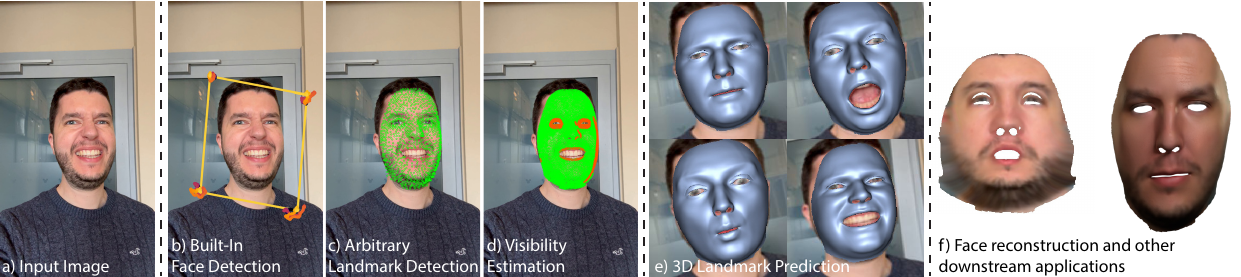}
   \centering
   \caption{Our Infinite 3D Landmark detector improves accuracy and temporal stability over existing detectors, is easy to use due to built-in face detection (b), can predict any number and layout of landmarks (c), and facilitates several downstream 3D applications like determining landmark visibility (d), 3D face reconstruction (e) and texturing (f).}
   \label{fig:teaser}
}

\maketitle
\begin{abstract}
   In this paper, we examine 3 important issues in the practical use of state-of-the-art facial landmark detectors and show how a combination of specific architectural modifications can directly improve their accuracy and temporal stability. First, many facial landmark detectors require a face normalization step as a preprocess, often accomplished by a separately-trained neural network that crops and resizes the face in the input image. There is no guarantee that this pre-trained network performs optimal face normalization for the task of landmark detection.  Thus, we instead analyze the use of a spatial transformer network that is trained alongside the landmark detector in an unsupervised manner, jointly learning an optimal face normalization and landmark detection by a single neural network. Second, we show that modifying the output head of the landmark predictor to infer landmarks in a canonical 3D space rather than directly in 2D can further improve accuracy.  To convert the predicted 3D landmarks into screen-space, we additionally predict the camera intrinsics and head pose from the input image. As a side benefit, this allows to predict the 3D face shape from a given image only using 2D landmarks as supervision, which is useful in determining landmark visibility among other things. Third, when training a landmark detector on multiple datasets at the same time, annotation inconsistencies across datasets forces the network to produce a suboptimal average. We propose to add a semantic correction network to address this issue. This additional lightweight neural network is trained alongside the landmark detector, without requiring any additional supervision. While the insights of this paper can be applied to most common landmark detectors, we specifically target a recently-proposed continuous 2D landmark detector to demonstrate how each of our additions leads to meaningful improvements over the state-of-the-art on standard benchmarks.
\end{abstract}



\section{Introduction}
\label{sec:intro}

Facial landmark detection is a well understood and heavily investigated problem in computer vision, with many applications in computer graphics.  For example, detecting a set of predefined 2D facial key points on an image is often an integral step in tasks like 3D face reconstruction, facial tracking, face image editing and deep face swapping.  There exists a plethora of algorithms for facial landmark detection, ranging from simple methods that rely on heuristics to deep neural networks trained on large annotated databases consisting of hundreds of thousands of images.  In this work, we look at three common issues plaguing many of the current state-of-the-art facial landmark detectors, and propose three extensions that, when combined, improve the practicality, accuracy and temporal stability of facial landmark detection.

First, most landmark detectors require a face normalization step as a preprocess, which is usually implemented as a separate pre-trained neural network that crops and resizes the face in the image.  In that case, the normalization process has no knowledge of the downstream landmark detection task, and as such there is no guarantee that the normalization network will create optimal input face images for landmark detection.  Furthermore, evidence shows that the normalized images can be temporally unstable - making the task more difficult for the landmark detector.  These issues can be alleviated by introducing a spatial transformer network that is trained alongside the landmark detector in an unsupervised manner, jointly learning an optimal face normalization and landmark detection at the same time by a single architecture.

Second, we show that landmark accuracy and stability can be improved by inferring the landmarks in a canonical 3D volume and projecting them onto a virtual camera plane to obtain the 2D landmark positions.  In this process we learn also the head pose and camera focal length from the input image.  Not only does this give more stable 2D landmarks, it also allows to predict the 3D shape of the face from the given image only using 2D landmarks as supervision.  Obtaining the landmarks in 3D also helps determine the visibility of the predicted landmark set, which is very useful for downstream applications.

Third, most deep landmark detectors are trained on multiple datasets from different sources at the same time, each dataset containing many face images and corresponding 2D landmark annotations.  Most datasets aim to portray the same semantic set of 68 landmarks on the face, facilitating cross-dataset training.  Unfortunately, due to inconsistencies in human annotation, there often exists a minor discrepancy in landmark semantics from one dataset to another.  As an illustrative example, consider a landmark at the tip of the nose.  In one dataset the annotations may be consistently higher than in another dataset, essentially corresponding to a different semantic location on the face.  Existing landmark detectors do not account for this and can thus result in a suboptimal averaging across datasets.  We propose to add a lightweight semantic correction network that can predict the per-dataset inconsistencies in semantics, resulting in overall higher accuracy when training on multiple datasets simultaneously.

While our work is not the first to use spatial transformers for face alignment or predict landmarks in 3-dimensions, the benefit of this paper is in exploring the effects of these architectural changes in modern facial landmark detectors.  Furthermore, to our knowledge this work is the first to propose a semantic correction network to improve training across datasets.  As we will propose, the best results are achieved with a novel combination of architecture changes.

Concretely, in order to demonstrate the three improvements we use a recently-proposed continuous 2D landmark detector~\cite{Chandran2023} as our baseline.  This baseline method represents the semantic landmarks as 3D query positions on a canonical face mesh, and the network takes a 3D query and a normalized face image as input and predicts the 2D pixel corresponding to the 3D query point.  The baseline method already shows state-of-the-art performance when predicting the standard set of 68 landmarks, and offers the additional feature that any additional points on the face may be predicted at runtime.  We therefore consider this a strong baseline to demonstrate our proposed architecture changes.
Despite the already strong performance of the baseline method, we will show that our proposed architecture changes improve accuracy and usability even further, advancing the state of the art in facial landmark detection as seen in \figref{fig:teaser}.

\section{Related Work}

As facial landmark prediction is one of the most studied fields in computer vision, doing a complete summary would be outside the scope of this paper.  Nevertheless, in the following we highlight the most relevant works with a focus on methods predicting 3D landmarks and state-of-the-art methods.  We refer the reader to the recent surveys of Wu~\etal~\cite{Wu2019}, Wang~\etal~\cite{Wang2014} and Khabarlak and Koriashkina~\cite{Khabarlak2022jcst} for a more in-depth review.

Traditionally, facial landmark detectors output 2D landmarks, corresponding to the locations on the image plane~\cite{Chandran2020HighResHG, wood2022dense, Chandran2023}.  Slightly less common, 3D landmark predictors have also been proposed in the literature.  For example, the Face Alignment Network from Bulat~\etal~\cite{Bulat2017} proposes an additional network that turns 2D landmark predictions into 3D, leveraging some 3D annotated data for training.  Zadeh~\etal~\cite{Zadeh_2017_ICCV} use a mixture of local convolutional experts network in an end-to-end framework, predicting 3D landmarks with heatmaps.  Yanda~\etal~\cite{Yanda2022} use a graph convolution network to predict 3D landmarks.  Another common way of including 3D knowledge when predicting landmarks is to leverage a generic 3D face model and deform it~\cite{Bhagavatula_2017_ICCV} or use an underlying 3D morphable face model~\cite{Bas_2017_ICCV,Zhu2016,guo2020towards}.  The method proposed by Basak~\etal~\cite{basak2023lightweight} predicts a denser set of 3D landmarks but can only output a fixed layout.

Recently introduced, Spatial Transformer Networks (STNs)~\cite{jaderberg2016spatial} are a class of neural networks that allow to spatially transform feature maps based on the feature maps itself, without additional supervision.  A very fitting application for these STNs is the prediction of a bounding box, or crop, used for a downstream application.  Some works use a STN to jointly learn a face alignment step with a face recognition network~\cite{Wu_2017_ICCV} or with a facial emotion recognition network~\cite{luna2021guided}.  More similar to us, Lv~\etal~\cite{Lv2017} use supervised Spatial Transformer Networks to re-initialize a pre-computed rough bounding box, they require additional supervision for the STN output while our proposed method can be trained in a fully unsupervised manner.

Recently, Wood et al. \cite{wood2022dense} proposed dense 2D facial landmark detection for 3D face reconstruction, and achieved state-of-the-art results by fitting a 3D morphable model to a dense set of around 700 facial landmarks \cite{wood2021fake}.  This work was succeeded by the work of Chandran et al. \cite{Chandran2023} that proposed a landmark detector capable of predicting an arbitrary number of facial landmarks ranging from arbitrarily sparse to arbitrarily dense layouts, thereby improving the flexibility of today’s facial landmark detectors.

Since our method also predicts 3D landmarks as an intermediate step, we also point out face reconstruction algorithms like TokenFace \cite{Zhang_2023_ICCV}, MICA \cite{MICA:ECCV2022}, DECA \cite{Feng:SIGGRAPH:2021}, and 3DDFAv2 \cite{guo2020towards}. Finally as we propose a query deformation network to address semantic inconsistencies in landmark annotations across multiple datasets, we also note the work of Meng \etal \shortcite{Meng2023} in dataset unification, which looked at aligning discrete object categories for object detection. However, their approach cannot be readily adapted to address semantic inconsistencies in continuous 2D landmark annotations as in our work.

In contrast to all previous methods, we believe our work is the first to combine a spatial transformer network for automatic bounding box detection in a continuous landmark detector that predicts 3D landmarks with a mechanism to account for annotation inconsistencies across training datasets.  Furthermore we both quantitatively and qualitatively demonstrate the benefits of our approach over a state-of-the-art method~\cite{Chandran2023}.

\section{Method}
\label{sec:method}

\begin{figure*}
    \begin{centering}
        \includegraphics[width=\textwidth]{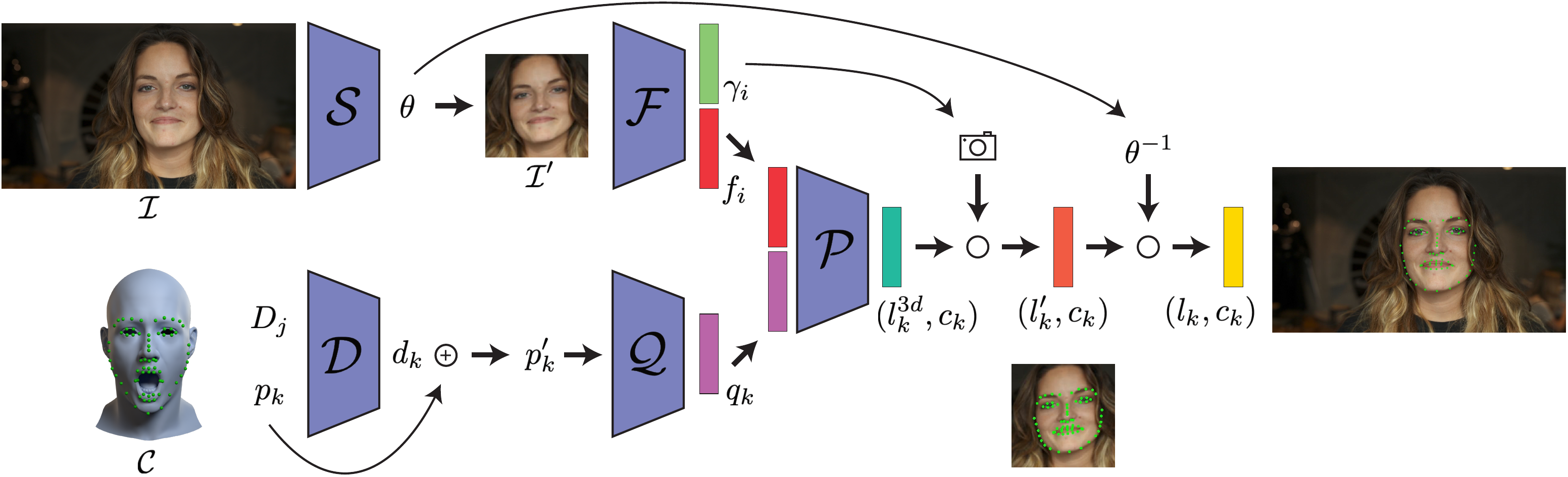}
        \caption{Our inputs include a face image $\mathcal{I}$ (un-normalized) and positions $p_k$ on a canonical shape $\mathcal{C}$.  A spatial transformer $\mathcal{S}$ auto-normalizes the face for the feature extractor $\mathcal{F}$, which predicts image features $f_i$ and camera plus head pose parameters $\gamma_i$.  Query points $p_k$ are passed through our new query deformer $\mathcal{D}$ to account for different datasets $D_j$, and are then position-encoded by $\mathcal{Q}$.  A 3D landmark predictor $\mathcal{P}$ estimates the landmarks in a canonical 3D space, which are projected to the camera plane and transformed back to original image space.}
        \label{fig:netArch}
    \end{centering}
\end{figure*}

We select the recently proposed continuous landmark detector of Chandran~\etal~\cite{Chandran2023} as our baseline architecture and make small yet impactful additions to it in our work. The continuous 2D landmark detector requires two inputs consisting of a normalized face image $\mathcal{I}'$ and query positions $p_k$ on a canonical 3D shape. Given these inputs, their method makes use of an image feature extraction network $\mathcal{F}$, and a queried landmark predictor $\mathcal{P}$ that predicts 2 outputs $(l_{k}, c_{k})$, where $l_{k}$ refers to the 2D coordinates and $c_{k}$ the confidence of each facial landmark that corresponds to a unique input query point, which is position-encoded using an MLP $\mathcal{Q}$. The networks $\mathcal{F}$, $\mathcal{P}$ and $\mathcal{Q}$ are trained end-to-end in a supervised manner using a collection of facial landmark datasets containing ground truth landmark positions. Chandran~\etal report multiple variations of their continuous 2D landmark detector, where each variant offers a tradeoff between prediction accuracy and speed. We use the
\emph{ConvNext + MLP} variant as our baseline architecture as it achieves a good balance between speed and accuracy. We refer to this architecture as the baseline method in the rest of the paper.

We will now describe the three extensions to this baseline, each of which notably improves the performance (as we will demonstrate in \secref{subsec:ablation}), while also adding new capabilities like landmark visibility estimation and face texture estimation to the network. Our extensions consist of i) A spatial transformer network $\mathcal{S}$ for built-in face localization (\secref{subsec:spatialTransformer}), ii) A modification of $\mathcal{P}$ to include a 3D landmark prediction head for improved overall accuracy (\secref{subsec:3dLandmarks}), and iii) A semantic correction network in the form of a query deformer network $\mathcal{D}$, designed to handle semantic inconsistencies in ground truth annotations across training datasets (\secref{subsec:queryDeformer}). Our final network architecture that includes all three extensions is shown in \figref{fig:netArch}.

\subsection{Spatial Transformers for Built-In Face Localization}
\label{subsec:spatialTransformer}
Facial landmark prediction algorithms, although differing in their details, often rely on detecting a bounding box of the face in the input image as a preprocessing step to simplify the job of the neural network (or algorithm). Despite the existence of several commonly used face detection techniques~\cite{dlibKazemi,MtcnnZhang}, this preprocessing step is often surprisingly susceptible to failures in practice, and often results in temporally unstable bounding boxes that can cause several problems for downstream applications (see \figref{fig:bboxTrajectory}). Furthermore as the bounding boxes for face detection were predetermined independent of landmark detection, there is also no guarantee that the such cropped faces are an optimal input to a landmark prediction network.

In our first addition to the baseline method, we introduce a Spatial Transformer Network \cite{jaderberg2016spatial} which replaces the explicit face detection and normalization step. Spatial Transformer Networks (STNs) were originally developed with the intention of offering a neural network the flexibility of geometrically transforming the input to maximize its training objective. STNs are typically small neural networks, designed to predict a parameterized 2D  transformation in an unsupervised manner, which is used to resample the input image before it is fed to a downstream neural network. Incorporating a spatial transformer network has indeed been explored previously in applications like face tracking, registration, recognition, etc~\cite{Bhagavatula_2017_ICCV,Wu_2017_ICCV,peebles2022gansupervised,TMPEH:CVPR:2023}. In our work, we revisit this idea and explore their applicability inside a state of the art continuous facial landmark detection system.

In \figref{fig:netArch}, we show how we introduce the spatial transformer $\mathcal{S}$ into the architecture of Chandran~\etal~\cite{Chandran2023}. Our spatial transformer is a convolutional neural network that takes the input image $\mathcal{I}$ and predicts the parameters $\theta$ of a 2D transform. A $2 \times 3$ transformation matrix is constructed from $\theta$ with which the input pixel grid is resampled to result in the normalized image $\mathcal{I}'$.

\begin{gather}
    \theta = \mathcal{S}(\mathcal{I}) \\
    \mathcal{I'} = \mathcal{W}(\mathcal{I};~\theta)
\end{gather}

Here $\mathcal{W}$ refers to a resampling operator that, given a transformation corresponding to $\theta$, resamples the original image $\mathcal{I}$ and provides the normalized image $\mathcal{I'}$.  The exact nature and number of parameters in $\theta$ depends on the class of the 2D transformation predicted by the spatial transformer. For example, a similarity transformation can be fully represented by 4 scalars that include an isotropic scale, a rotation in the image plane, and a 2D translation. On the other hand, 6 scalars are required to properly represent an affine transformation as it also models anisotropic scaling, shearing and so on. While any class of 2D transformations can be predicted by a spatial transformer, in our work we explored both similarity and affine transformations (see \secref{subsec:evalFaceNormalization}) and empirically found affine transformations to provide the best performance.

The warped image $\mathcal{I}'$ is the equivalent of the localized face image that is usually obtained using face detectors or other normalization techniques. The image  $\mathcal{I}'$ is then fed as input to the image feature extraction network $\mathcal{F}$ from Chandran \etal \cite{Chandran2023}. Different from the baseline method, the resulting 2D landmarks $l_{k}'$ lie in the screen space of $\mathcal{I'}$ and not $\mathcal{I}$. However the ground truth landmarks are still defined with respect to the original image $\mathcal{I}$. Therefore we restore the predicted landmarks $l_{k}'$ to the original image $\mathcal{I}$ using the inverse spatial transformation corresponding to $\theta^{-1}$ to result in $l_{k}$.
\begin{equation}
    \label{eqn:inverseTransformLandmarks}
    l_k = \mathcal{T}(l_{k}';~\theta^{-1}),
\end{equation}
where $\mathcal{T}$ denotes applying the 2D transformation corresponding to $\theta^{-1}$ on the landmarks $l_{k}'$.
The spatial transformer network is trained alongside the rest of the network in an end-to-end fashion.  As the output of the spatial transformer is unsupervised, it is free to learn any transformation of the input such that landmark prediction loss is minimized. In \secref{subsec:evalFaceNormalization} we discuss some interesting properties of learning unsupervised face RoIs with a spatial transformer.

\subsection{3D Landmark Prediction}
\label{subsec:3dLandmarks}
Our second extension is a reformulation of how the 2D landmarks are predicted. In the baseline architecture, the output of the queried landmark predictor $\mathcal{P}$ are  normalized 2D landmarks $l_{k}$ $\in$ [-1, 1], and a 1D confidence $c_{k}$ value for each landmark. When running landmark detection on in-the-wild videos, as a persons speaks and moves around, certain landmarks become occluded; like the jawline landmarks as a person turns to one side. Having knowledge of the visibility of the predicted landmarks can be valuable for downstream applications like 3D face reconstruction, giving a reconstruction algorithm the ability to trust invisible landmarks less. While the confidence values $c_{k}$ predicted by the baseline method have some correlation to visibility, the confidence values are not always semantically interpretable and therefore are not guaranteed reason about landmark visibility. To address this problem, we modify the queried landmark predictor $\mathcal{P}$ such that it predicts 3D landmarks in a canonical space, which are posed and reprojected onto the screen to obtain 2D landmarks. As a result, our method is able to accurately reason about the visibility of the predicted 2D landmarks. Furthermore, in our approach the 3D landmarks are learned in an unsupervised manner and are predicted in a continuous fashion for each input query. This allows users to predict virtually an unlimited number of 2D landmarks in any configuration, and reason about their visibility. We will next describe the details of our 3D landmark prediction approach.

As human faces deform in a characteristic way, facial landmarks are often strongly correlated with one another. Applications such a 3D face reconstruction \cite{Feng:SIGGRAPH:2021} try to leverage this fact by making use of a 3D shape prior in the form of a morphable face model to predict plausible face shapes even in challenging, less constrained scenarios. These 3D priors play an important role in mitigating failure cases and always producing reasonable face-like outputs. We incorporate such an increased robustness into continuous 2D landmark detection without requiring a morphable model, by predicting 3D landmarks as offsets on top of a mean face shape $\mathcal{M}$. To accommodate this extension, we make 2 changes to the baseline's image feature encoder $\mathcal{F}$ and the queried landmark predictor $\mathcal{P}$ which are described below.

\paragraph*{Head Pose and Camera Estimation.}
We modify the image feature encoder $\mathcal{F}$ such that when given a normalized image $\mathcal{I'}$, in addition to predicting the image feature descriptor ${f}_i$, it predicts a vector $\gamma_i$ consisting of head pose $(R, T)$ and camera intrinsics $(f_{d})$.

\begin{gather}
    f_{i}, \gamma_{i} = \mathcal{F}(\mathcal{I'})\\
    \gamma_{i} = [R, T, f_{d}]
\end{gather}

We parameterize head pose as a 9D vector consisting of a 6D rotation vector $R$~\cite{Zhou2019} and a 3D translation $T$. While in theory only the head pose is enough to re-project a 3D landmark through a fixed canonical camera, we empirically found that predicting camera intrinsics allows for increased accuracy (see \secref{subsec:ablation}). For the camera intrinsics, we only predict a single focal length in millimeters (mm) under an ideal pinhole assumption. To bias the training towards plausible focal lengths, the focal length in $\gamma_i$ is a focal length displacement $f_{d}$ that is added to a predefined focal length $f_{fixed}$ which we set to 60mm.

\paragraph*{Unsupervised 3D Landmark Prediction.}
The queried landmark predictor $\mathcal{P}$ in the baseline architecture predicts a 3-dimensional output $(l_{k}, c_{k})$. We increase the dimensionality of the output to 4 dimensions such that it now predicts $(l_{k}^{3d}, c_{k})$, where $l_{k}^{3d}$ is a 3D offset vector. For each query point $p_{k}$, $\mathcal{P}$ predicts a 3D offset that is added to the corresponding point on a mean face shape $m_{k}^{3d}$ to result in the canonical 3D position $L_{k}^{3d}$ of the queried landmark.

\begin{gather}
    q_{k} = \mathcal{Q}(p_{k}') \\
    (l_{k}^{3d}, c_{k}) = \mathcal{P}(f_{i}, q_{k}) \\
    L_{k}^{3d} = l_{k}^{3d} + m_{k}^{3d}
\end{gather}

Here, $p_{k}'$ is the deformed query point (described next in \secref{subsec:queryDeformer}) and $\mathcal{Q}$ is a position-encoding MLP.  Note that there are no special requirements on the mean face shape $m_{k}^{3d}$, other than that we recommend it shares the same topology as the canonical face shape $\mathcal{C}$ for ease of use. The canonical 3D position $L_{k}^{3d}$ of a landmark is then transformed using the head pose $(R, T)$ predicted by the image feature encoder $\mathcal{F}$ to result in $\bar{L}_{k}^{3d}$. Then $\bar{L}_{k}^{3d}$ is projected through a canonical camera with a focal length of $f_{fix} + f_{d}$ to result in the normalized 2D landmark $l_{k}'$.

\begin{gather}
    \bar{L}_{k}^{3d} = \mathcal{T}(L_{k}^{3d};~R, T) \\
    l_{k}' = \psi(\bar{L}_{k}^{3d};~f_{fixed} + f_{d})
\end{gather}

These normalized landmarks $l_{k}'$ are restored to the screen space of the input image $\mathcal{I}$ using $\theta^{-1}$ resulting in the final 2D landmarks $l_{k}$. The confidence values $c_{k}$ of the 3D landmarks $L_{k}^{3d}$ are transferred over to the 2D landmarks $l_{k}$ for training with the Gaussian NLL loss. This allows our approach to infer 3D landmarks while continuing to supervise all networks with only 2D ground truth as before.

\subsection{Query Deformer for Inconsistent Landmark Annotations}
\label{subsec:queryDeformer}

The third and final contribution of our work addresses the practical issue of training a facial landmark detector on multiple datasets simultaneously. While most datasets aim for consistent annotations within the dataset, it can be the case that different datasets are slightly inconsistent across the datasets, even for the same semantic landmarks on the face.  Additionally, our baseline method from Chandran~\etal~\cite{Chandran2023} has the added benefit that it can be trained on datasets with vastly different landmark configurations.  However one drawback of their approach is the reliance on per-dataset queries that have to be predefined or hand annotated on the canonical shape $\mathcal{C}$.  This is another source of potential annotation mistakes, leading to additional inconsistencies.

While post process strategies like label translation \cite{wood2022dense} and query optimization \cite{Chandran2023} do alleviate this problem to some extent, they are only mitigation strategies and do not address the problem directly.

In our work, we tackle this problem at training time by proposing a query deformer module $\mathcal{D}$. Given a query point $p_{k}$ and a dataset identifier $D_{j} \in \mathcal{R}^N$ the query deformer predicts a displacement $d_{k}$ of the query point. The displacement $d_{k}$ is added to the input query $p_{k}$ to result in a canonical query $p_{k}'$ that is learned during training to represent all datasets fairly.  However when using a query deformer module, it is important to ensure that queries corresponding to different datasets continue to remain on the manifold of the canonical face $\mathcal{C}$. To ensure this, we operate in the parametric UV space of the canonical face and provide 2D UV queries as input to the query deformer $\mathcal{D}$, resulting in 2D displacements. The displaced UV coordinate is used to sample a position map of the canonical face to result in the 3D query. This 3D query, $p_{k}'$ is then fed as input to the rest of the pipeline as shown in \figref{fig:netArch}. With this modification, our new continuous landmark predictor has the option of deforming queries $p_{k}$ from the training datasets to a different position on the canonical shape such that inconsistent query annotations for the same semantic landmark across datasets can be corrected for during training.

The dataset code $D_{j}$ is a N dimensional vector per training dataset that is optimized for along with the training of the landmark predictor. For example, when training with the studio dataset of Chandran \etal \cite{Chandran2020sdfm} together with a synthetic dataset such as the one of Wood \etal \cite{wood2021fake}, we set N=2 and optimize for two different codes $D_0$ and $D_1$.

With these 3 extensions to the baseline architecture of Chandran \etal~ \cite{Chandran2023}, we are able to predict an unlimited number of 2D/3D landmarks on face images without requiring an explicit face detection step, and while also accommodating inconsistencies in annotations across training datasets. With these modifications in mind, we refer to our new landmark detector as the \emph{Infinite 3D Landmark Predictor} while evaluating our results in \secref{sec:results}.

\section{Implementation Details}
\label{sec:training}

\textbf{Training Data}. For training we exclusively use a studio dataset consisting of dense facial skin landmarks \cite{Chandran2020sdfm} and an in-the-wild synthetic dataset containing sparse landmark annotations \cite{wood2021fake}. We annotate queries on the canonical shape for both datasets similar to Chandran \etal \cite{Chandran2023} and train both the baseline method and our proposed extension from scratch. In total, our training dataset consists of 37,344 studio images with 47,022 dense facial skin landmarks and 100,000 in-the-wild images with 68 sparse facial landmarks. Our evaluation data is the common subset of Sagonas~\etal~\cite{Sagonas300Faces} and contains 554 images.

We perform various photometric and geometric augmentations on the training images and landmarks to increase the generalization capabilities of our network. We train all reported methods for 25 epochs, with a batch size of 64 on an A6000 GPU. We use a the AdamW \cite{Loshchilov2017} optimizer with a learning rate of 1e-4.

\noindent\textbf{Network Details}. We used the \emph{convnext\_tiny} model for our spatial transformer network $\mathcal{S}$, and replaced the last linear layer to predict 4 and 6 outputs for the similarity and affine STN respectively. Similar to the baseline method of Chandran \etal \shortcite{Chandran2023}, we use the \emph{convnext\_base} model for our feature extraction network $\mathcal{F}$. The query deformer $\mathcal{D}$, and the position encoder $\mathcal{Q}$, are MLPs with 2 hidden layers with GeLU activations and contain 64 neurons per hidden layer. The landmark prediction MLP $\mathcal{P}$, consists of 4 hidden layers with 512 neurons per layer. All networks were written using standard building blocks available in the \texttt{torch} and \nobreak\texttt{torchvision} packages.

\noindent\textbf{Runtime}. As we use small networks for both the spatial transformer and the query deformer networks, our work adds minimal overhead in terms of computation time to the baseline method. Our final network consisting of all proposed components runs at 46 fps on a RTX 3090 GPU in comparison to the baseline method which runs at 48 fps.
\vspace{-3mm}
\section{Results}
\label{sec:results}
We now showcase various results of our method, and evaluate how each of our design choices improves the overall performance over the baseline method \cite{Chandran2023}.

\subsection{Qualitative Results}

\begin{figure*}
    \begin{centering}
        \includegraphics[width=\textwidth]{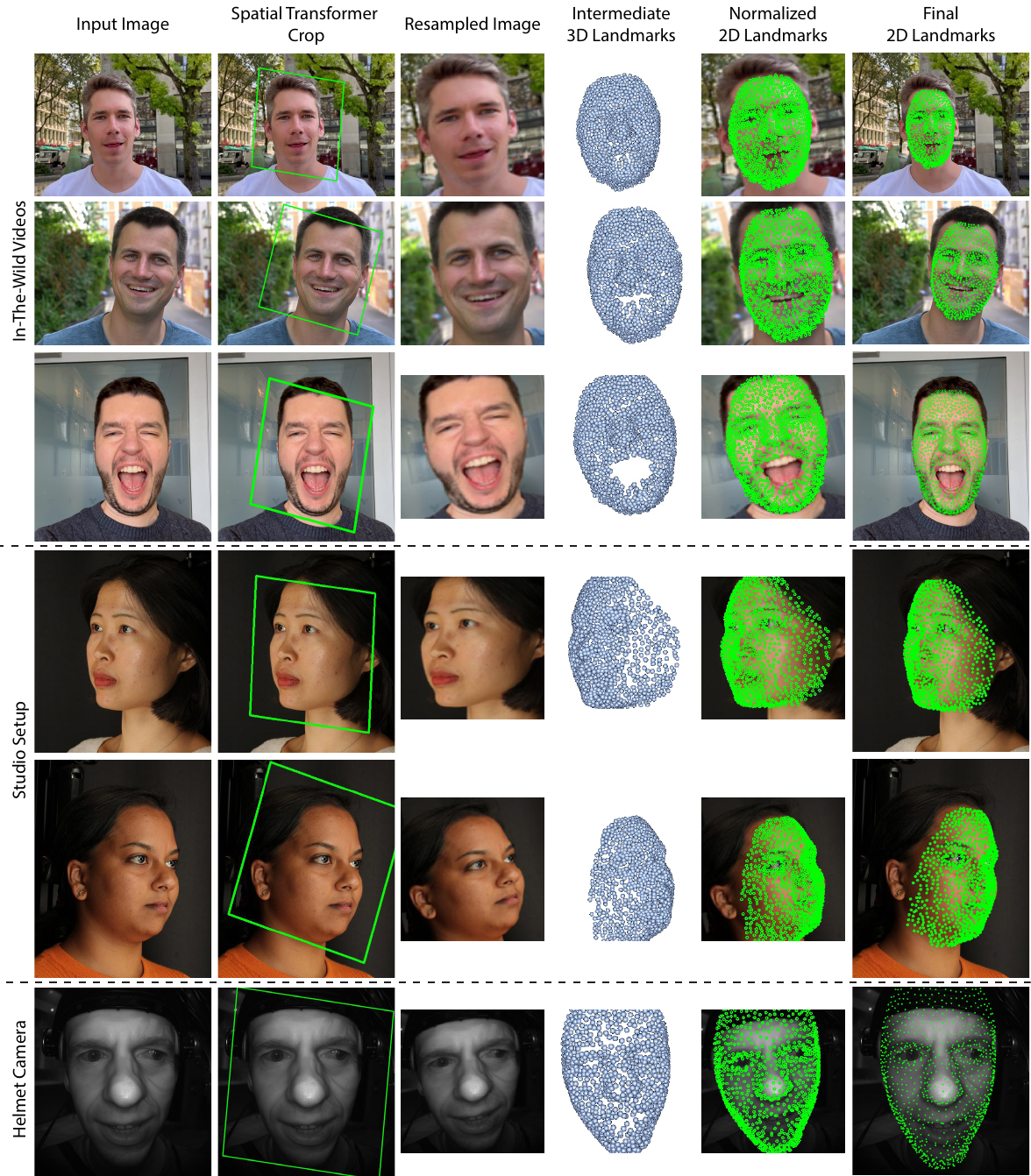}
        \caption{Our method can predict accurate facial landmarks on a number of practical scenarios including studio setups, in-the-wild videos, mobile phone recordings, and even helmet mounted cameras. We show the result of each stage of our pipeline with the input image $\mathcal{I}$ (first column), the RoI detected by the spatial transformer (second column), the resampled or normalized face image $\mathcal{I}'$ (third column), the intermediate 3D landmarks predicted by the model $\bar{L}_{k}^{3d}$ (fourth column), the resulting 2D landmarks ${l}_{k}'$ corresponding to $\mathcal{I}'$ (fifth column), and the final landmark positions $l_{k}$ (last column).}
        \label{fig:heroResults}
    \end{centering}
\end{figure*}

In \figref{fig:heroResults} we show the stagewise results of our landmark detection pipeline on images captured under various practical scenarios including in-the-wild videos, multiview studio setups, and helmet mounted cameras. We show results for both portrait and landscape aspect ratios and for resolutions ranging from 256 x 256 in-the-wild images to 4K resolution studio setups. In all cases, similar to the majority of existing facial landmark predictors, the images are square padded and resized to 256 x 256 before feeding them as input to our network.
Our jointly trained spatial transformer network is able to localize the face consistently in all scenarios as shown in the second column. Resampling the image with the output of the spatial transformer results in normalized face images $\mathcal{I}'$ (third column). These images are fed to the image feature encoder $F$ and the rest of our landmark prediction pipeline producing intermediate 3D landmarks $\bar{L}_{k}^{3d}$ (fourth column), and ultimately the final 2D landmarks $l_{k}$ (final column). As illustrated in this figure, our method provides good results for all of these complex scenarios.

\begin{figure}
    \begin{centering}
        \includegraphics[width=\columnwidth]{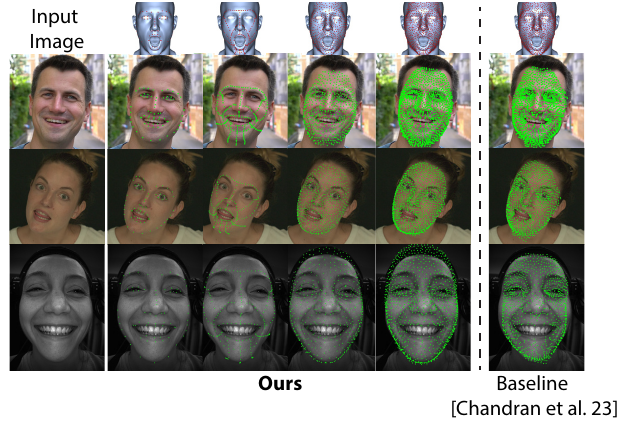}
        \caption{While our method remains competitive with the state-of-the-art baseline in common scenarios (first two rows), it provides significantly better results on challenging scenarios like helmet mounted cameras, where our method is able to capture the overall head shape and expression better than the baseline (last row). Queries $p_{k}$ corresponding to each landmark layout are visualized at the top.}
        \label{fig:qualCompare}
    \end{centering}
\end{figure}

As our method improves on the continuous landmark detection of Chandran \etal~ \cite{Chandran2023}, we are able to predict an unlimited number of 2D landmarks in any configuration on arbitrary images of human faces. In \figref{fig:qualCompare}, we show a qualitative comparison of 2D landmarks on in-the-wild videos versus the baseline method of Chandran \etal \cite{Chandran2023}. Our method retains the flexible nature of the baseline in predicting continuous, arbitrary facial landmarks under, while additionally not requiring face normalization as a pre-processing step. While both methods show comparable performance on common in-the-wild videos, our method starts to significantly outperform the baseline on challenging test conditions like helmet camera data as seen in the last row. The landmarks predicted by our method capture the overall face shape and expression better than Chandran \etal when trained on the same data. We hypothesize that our 3D landmark prediction, which makes use of a mean face shape and our estimation of camera instrinsics, jointly help make our method more robust than the baseline.

\subsection{Quantitative Evaluation}
We now discuss quantitative evaluation of our infinite 3D landmark predictor on a popular 2D facial landmark benchmark \cite{Sagonas300Faces} in \tabref{table:ablation}. Conventionally this benchmark provides both training and test data to compare landmark prediction algorithms. However we find that the training data in this benchmark contains copyrighted images and so to respect copyright, we exclusively use only the test data from \cite{Sagonas300Faces} for evaluation and do not finetune our networks on the training data. For quantitative evaluation, we report the Normalized Mean Error (NME) in \tabref{table:benchmark}. As it would practically be infeasible to retrain previous landmarks detectors on the same data as what we use in our evaluation, we leave out other state of art methods from this table to avoid confusion. Finally our training data (see \secref{sec:training}) consisting of studio \cite{Chandran2020sdfm} and synthetic data \cite{wood2021fake} contains 3D consistent landmarks even for occluded points like the jawline, and having been trained on this data, our method always predicts 3D consistent landmarks. As the benchmark evaluation data contains sliding landmarks on the jawline that are not 3D consistent, we leave out the 17 jaw landmarks from the ground truth test data and only use the remaining 51 landmarks for quantitative evaluation. This avoids the need to perform mitigation strategies like label translation \cite{wood2022dense} and enables us to fairly showcase the magnitude of our improvements.

In addition to reporting the spatial accuracy of the landmarks, we compare the temporal stability of predicted landmarks on left out dynamic sequences from a studio dataset \cite{Chandran2020sdfm} where perfect ground truth is available. We report a temporal normalized mean error in \tabref{table:temporal}. This temporal metric is computed as follows

\begin{equation}
    E_{temporal} = \frac{1}{NT}\sum_{t=1}^{T}\sum_{k=1}^{N}\frac{||p_{k}^{t + 1} - p_{k}^{t}||_{2} - ||g_{k}^{t + 1} - g{k}^{t}||_{2}}{||g_{k}^{t + 1} - g{k}^{t}||_{2}}
\end{equation}
where $p_{k}^{t}$ and $g_{k}^{t}$ refer respectively to the $k^{th}$ predicted and ground truth landmarks in frame $t$.  This temporal metric ignores absolute positional errors between the predicted and ground truth landmarks and only concerns itself with the average difference in velocities of a predicted landmarks in subsequent frames with respect to the corresponding landmarks in the ground truth.

Our method outperforms the state-of-the-art baseline of Chandran \etal \cite{Chandran2023} on both metrics, thereby quantitatively corroborating the value of our contributions.

\begin{table}[t]
    \centering \small
    \caption{Quantitative Evaluation on the Common Benchmark of Sagonas~\etal~\cite{Sagonas300Faces}}
    \begin{tabular}{|l|c|}
        \hline
        \textbf{Method}                               & \textbf{NME}  \\
        \hline
        Baseline (ConvNext + MLP) \cite{Chandran2023} & 3.19          \\
        \hline
        Ours                                          & \textbf{2.89} \\
        \hline
    \end{tabular}
    \label{table:benchmark}
\end{table}

\begin{table}[t]
    \centering \small
    \caption{Temporal Stability Metric computed on Test Studio Videos}
    \begin{tabular}{|l|c|}
        \hline
        \textbf{Method}                         & \textbf{Temporal Error} \\
        \hline
        Baseline \cite{Chandran2023}            & 3.05                    \\
        \hline
        Baseline + Spatial Transformer (Affine) & 2.41                    \\
        \hline
        Baseline + 3D landmarks                 & 3.05                    \\
        \hline
        Ours                                    & \textbf{2.37}           \\
        \hline
    \end{tabular}
    \label{table:temporal}
\end{table}

\subsection{Evaluation}
\label{sec:evaluation}
Having demonstrated the qualitative and quantitative superiority of our method over state of the art, in this section we take a closer look at the three extensions to baseline method: i) the spatial transformer network, ii) the 3D landmark prediction, and iii) the query deformation module. We also discuss how each of them contribute to the overall performance of the system in \secref{subsec:ablation}.

\subsubsection{Face normalization with the spatial transformer}
\label{subsec:evalFaceNormalization}
We look at face normalization as performed by the spatial transformer. As seen in the second column of \figref{fig:heroResults}, our affine spatial transformer can handle images of different aspect ratios and can consistently localize the face in several scenarios not restricted to in-the-wild videos, studio capture sessions, and helmet mounted cameras.

Conventionally faces are detected using a dedicated face detection algorithm such as Kazemi \etal~\cite{dlibKazemi} or a more recent neural approach of Zhang \etal~\cite{MtcnnZhang}. As our method removes this explicit face detection step with a spatial transformer network, we compare the facial bounding boxes predicted by a state of the art bounding box detector againt the learned region of interest (RoIs) predicted by the spatial transformer.

\begin{figure}
    \begin{centering}
        \includegraphics[width=\columnwidth]{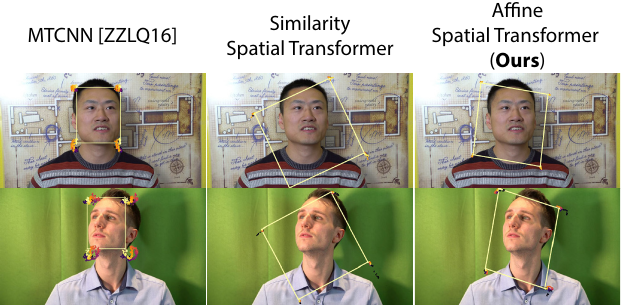}
        \caption{We visualize the bounding box trajectories on test videos. In the first column, we show predictions from the widely used face detection algorithm of Zhang \etal. While predicting a tighter crop of the face, the method of Zhang \etal results in a noisy trajectory for the bounding box even with very little movement of the face. The learned RoIs predicted by both the similarity (second column) and affine (third column) spatial transformers, while larger in frame, are temporally smoother.}
        \label{fig:bboxTrajectory}
    \end{centering}
\end{figure}

In \figref{fig:bboxTrajectory} we show a qualitative comparison of the bounding box and their trajectories on typical in-the-wild test videos. We compare two variants of our spatial transformer; each of which predicts a similarity and an affine transformation respectively against the widely used method of Zhang \etal~ \cite{MtcnnZhang}. From the displayed trajectories, it is evident that a jointly trained spatial transformer predicts temporally smoother bounding boxes when compared to the method of Zhang \etal. We kindly refer you to our supplemental video for a better demonstration of the temporal smoothness of our learned bounding boxes. While both the similarity and the affine spatial transformer produce comparably smooth bounding box trajectories, the affine spatial transformer obtained a better score on our ablation study (see \secref{subsec:ablation}).

The second interesting inference from \figref{fig:bboxTrajectory} is that irrespective of the class of the transformation it predicts, the spatial transformer always prefers slightly rotated bounding boxes that place the face more or less along the diagonal of the bounding box. Currently we do not have an explanation for this preference and we find it an intriguing phenomenon.

\noindent\textbf{Spatial Transformers on Convolutional Architectures.}
Finally as spatial transformer networks can also be integrated as a standalone component into other convolutional architectures that are commonly used in keypoint detection, we present an evaluation where we prepend a spatial transformer to an hourglass network \cite{newell2016stacked} and measure the improvement it provides in facial landmark detection. To support end-to-end training with a heatmap regression network, we use the \emph{softargmax} operator to convert heatmaps into 2D landmark coordinates in normalized image space \cite{Chandran2020HighResHG}, and restore the normalized landmarks to the original input space by inverting the transformation predicted by the spatial transformer (see \eqnref{eqn:inverseTransformLandmarks}). When the hourglass network is trained in such a manner on a synthetic dataset \cite{wood2021fake}, and evaluated on the 300-W common benchmark, adding the spatial transformer lowers the NME of the hourglass network from \textbf{5.15} to \textbf{4.91}. This demonstrates that our end-to-end training strategy with a spatial transformer has benefits that go beyond the continuous landmark detection framework that we use as a baseline in our work.

\subsubsection{Infinite 3D Facial Landmarks}
\label{subsec:eval3DLandmarks}
\begin{figure}
    \begin{centering}
        \includegraphics[width=\columnwidth]{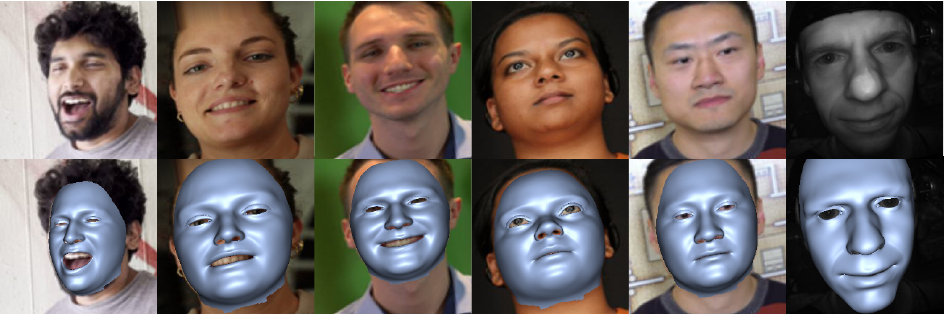}
        \caption{We visualize the normalized image $\mathcal{I'}$in the first row and an overlay of a mesh created using $\bar{L}_{k}^{3d}$ on the image in the second row. The tight overlay of the mesh on the image demonstrate the strong performance of unsupervised pose estimation from $\mathcal{F}$ and 3D landmark predictor $\mathcal{P}$.}
        \label{fig:faceMeshOverlay}
    \end{centering}
\end{figure}

\paragraph*{Face Reconstruction.} Our new 3D landmark predictor extends the method of Chandran \etal by predicting an arbitrary number of 3D facial landmarks in any layout on normalized input image $\mathcal{I}'$. Contrary to most existing 3D face reconstruction method, our 3D landmark predictor only predicts the 3D points corresponding to the input queries $p_k$. However by densely querying every point on the canonical shape $\mathcal{C}$, our method can readily be used to provide a full face mesh that matches $\mathcal{I'}$. In \figref{fig:faceMeshOverlay}, we visualize the mesh obtained by densely querying our landmark predictor and overlay it on the normalized image $\mathcal{I'}$. The results indicate that we learn plausible 3D face shapes for in-the-wild images even though we only use sparse 2D landmark supervision for training.

\begin{figure}
    \begin{centering}
        \includegraphics[width=\columnwidth]{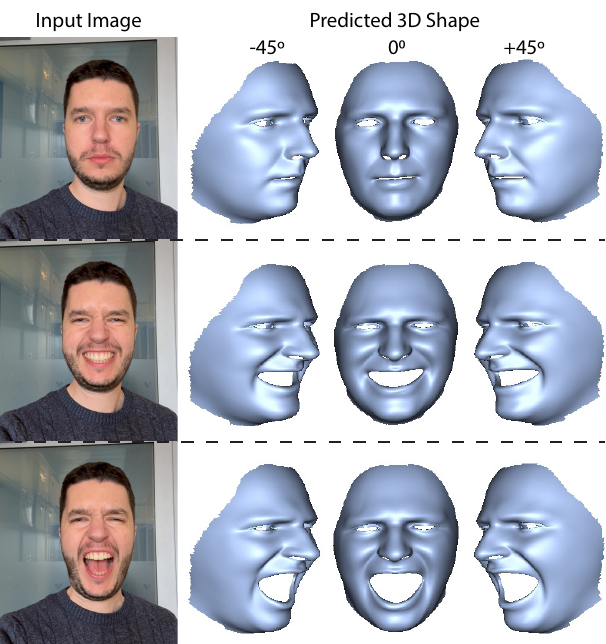}
        \caption{We visualize the predicted canonical shape from 3 different views (2 profile and 1 frontal) to demonstrate that our method can predict plausible facial geometry even for extreme expressions.}
        \label{fig:meshSideView}
    \end{centering}
\end{figure}

\noindent Even in the absence of constraining the predicted vertex offsets with a shape prior; such as a 3DMM, our method produces plausible face shapes. In \figref{fig:meshSideView}, we show several examples of the predicted canonical face shape for an input image from multiple views. The predicted canonical shape looks plausible even under extreme expressions. We hypothesize that this could be a consequence of the dense 2D supervision from the studio dataset.

\begin{figure}
    \begin{centering}
        \includegraphics[width=\columnwidth]{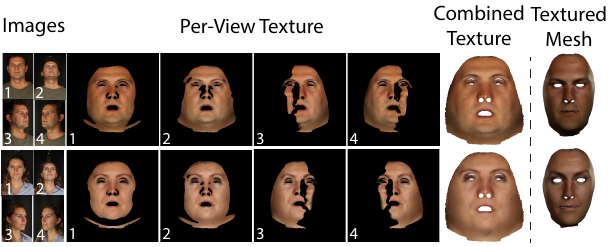}
        \caption{Our ability to predict 3D landmarks allows for the recovery of a full face texture from multiview images. We show images captured in a studio setting from 4 different viewpoints in column 1. Columns 2-5 show the view-specific texture map reconstructed by the dense prediction of 3D landmarks on the input images. The number corresponding to the view from which the texture  was reconstructed is shown in the bottom left of the per-view textures. Column 6 shows the combined texture spanning the full face that is obtained by averaging the per-view textures. In the last column, we apply the texture on the predicted 3D face mesh for visualization.}
        \label{fig:texCompletion}
    \end{centering}
    \vspace{-7mm}
\end{figure}

\paragraph*{Texture Completion.} One useful application of having the ability to implictly predict a mesh in the topology of the canonical shape is that it allows us to recover the texture of person's face from multiview images or a video. In \figref{fig:texCompletion}, we show the recovery of a full face texture from a multiview studio setup consisting of 4 cameras. For each view, we first pass the image through our landmark detector and predict 3D landmarks corresponding to each skin point on the canonical face mesh. Then we reproject the RGB colors from the normalized image $\mathcal{I'}$ onto the posed mesh that is created using $\bar{L}_{k}^{3d}$ and share the same triangles as $\mathcal{C}$. The reprojected colors are unwrapped into a texture using the UV parameterization of the canonical face $\mathcal{C}$, allowing us to create view specific textures for each input. These textures are then merged to a single combined texture (by averaging across the views).

\begin{figure*}
    \begin{centering}
        \includegraphics[width=\textwidth]{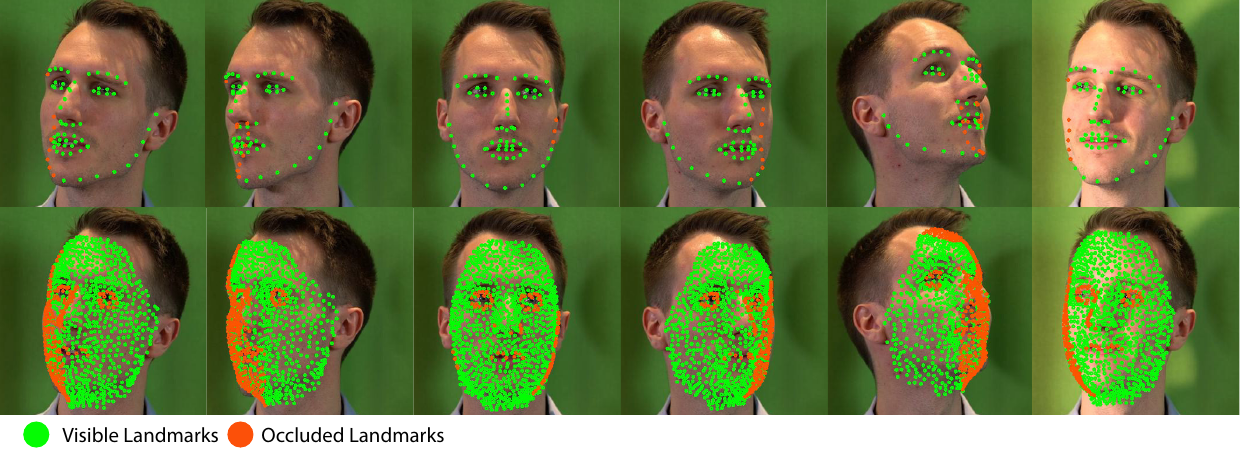}
        \caption{Our method can result in accurate visibility labels for any desired facial landmark. We show the visibilities estimated on a video for two different landmark layouts in rows 1 and 2. Our visibility labels accurately reflect the motion of the subject's head.}
        \label{fig:visibility}
    \end{centering}
\end{figure*}

\paragraph*{Visibility Estimation.} Our method thus has the ability to predict arbitrary 2D landmarks on the image, and to produce a dense 3D face mesh that can overlay well on the normalized image $\mathcal{I}$'. As a consequence of both of these abilities, we can accurately estimate the visibility of arbitary 2D facial landmarks on an image. In \figref{fig:visibility}, we demonstrate this new capability of visibility estimation that our method adds to the baseline approach. These landmark visibility estimates can later be used by downstream applications (like 3D face reconstruction) to assign weights to different landmarks based on their visibility.

\begin{figure}
    \begin{centering}
        \includegraphics[width=\columnwidth]{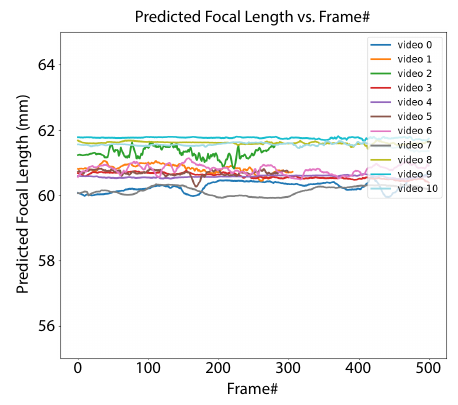}
        \caption{We visualize the estimated focal lengths on 11 test videos and find that they remain reasonably consistent and stay within an acceptable range.}
        \label{fig:focusStability}
    \end{centering}
\end{figure}

Lastly though our approach produces temporally smoother results than the baseline as seen in \tabref{table:temporal}, our method still operates on each image independently. Therefore while processing videos, our image encoder $\mathcal{F}$ can estimate slightly different focal length displacements $f_{d}$ for each frame in the video as we impose no other constraints on the training of the landmark detector other than 2D landmark supervision. We visualize the predicted focal lengths on test videos at inference time in \figref{fig:focusStability} and found that although the focal length changes across the video, the values stay reasonably consistent and always in a meaningful range.

\subsubsection{Query Deformation Module}
\label{subsec:evalQueryCorrection}
\begin{figure}
    \begin{centering}
        \includegraphics[width=\columnwidth]{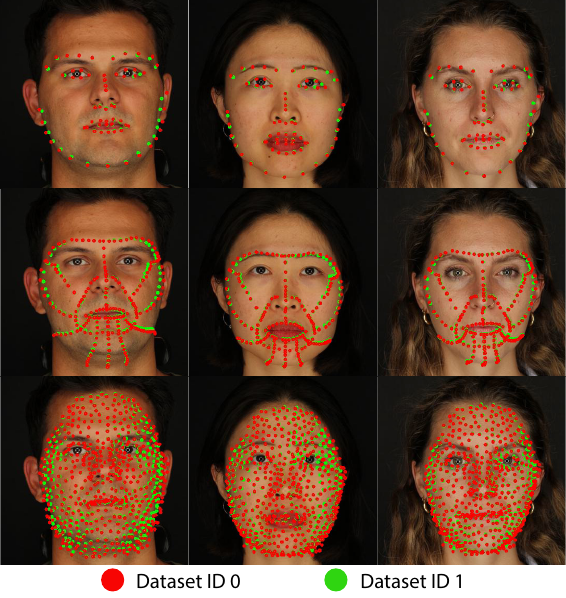}
        \caption{We visualize the effect of varying the dataset ID at inference time and how this shifts the landmarks slightly reflecting the original styles in which the two different training datasets were annotated.}
        \label{fig:datasetId}
    \end{centering}
\end{figure}

\paragraph*{Controlling Landmark Styles.}
The query deformation module, while allowing the network to account for query inconsistencies across datasets at training time, also allows us to infer the same landmarks in different styles at inference time by varying the dataset ID $D_{j}$. In \figref{fig:datasetId}, we show the same set of facial landmarks predicted on a test image, when using two learned dataset codes $D_{0}$ and $D_{1}$ which correspond to the studio dataset of Chandran \etal \cite{Chandran2020sdfm} and the synthetic dataset of Wood \etal respectively.

\subsection{Ablation Study}
\label{subsec:ablation}
Finally we quantitatively measure the effect of each of our additions to the baseline method of Chandran \etal~using the  normalized mean error (NME) on the benchmark of Sagonas \etal \cite{Sagonas300Faces}. Adding the spatial transformer, the 3D landmark predictor, and the query deformation module to the baseline architecture individually improves the performance of the baseline method. Our infinite 3D landmark detector includes the best performing variations all of our 3 proposed extensions and consists of an Affine Spatial Transformer, 3D landmark prediction with focal length displacements $f_{d}$, and the query deformation module. We clarify that while reporting the NME when using the query deformation module, we use a dataset code that resulted in the lowest error, which corresponded to the code of the studio training dataset \cite{Chandran2020sdfm}.

\begin{table}[t]
    \centering \small
    \caption{Quantitative Evaluation on the 300-W Benchmark}
    \begin{tabular}{|l|p{1cm}|p{1.5cm}|}
        \hline
        \textbf{Method}                                   & \textbf{Common Set} & \textbf{Challenging Set} \\
        \hline
        Baseline \textbf{B} \cite{Chandran2023}           & 3.19                & 6.37                     \\
        \hline
        \textbf{B} + Spatial Transformer (Similarity)     & 3.13                & 6.29                     \\
        \hline
        \textbf{B} + Spatial Transformer (Affine)         & 3.07                & 6.22                     \\
        \hline
        \textbf{B} + 3D landmarks ($f_{fixed}$)           & 2.99                & 5.75                     \\
        \hline
        \textbf{B} + 3D landmarks ($f_{fixed}$ + $f_{d}$) & 2.96                & 5.76                     \\
        \hline
        \textbf{B} + Query Deformation                    & 3.13                & 5.93                     \\
        \hline
        Ours                                              & \textbf{2.89}       & \textbf{5.71}            \\
        \hline
    \end{tabular}
    \label{table:ablation}
\end{table}

\section{Limitations And Failure Cases}
\begin{figure}
    \begin{centering}
        \includegraphics[width=\columnwidth]{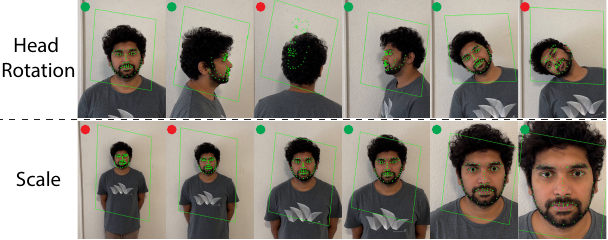}
        \caption{We show examples of failure cases (highlighted by red dots) involving strong head rotations (first row) and large changes in scale (second row). While our method can produce reasonable predictions for profile views, it starts to break down as the subject turns around completely. Strong in-plane rotations of the head are also a challenging case. Our method gracefully degrades in quality as the scale of the face in the input image changes drastically.}
        \label{fig:failureCases}
    \end{centering}
\end{figure}

\begin{figure}
    \begin{centering}
        \includegraphics[width=\columnwidth]{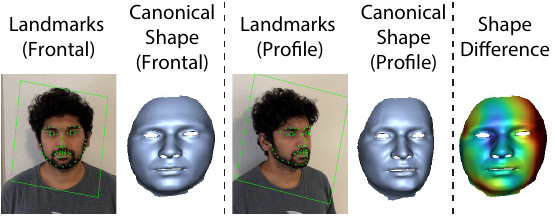}
        \caption{Even when the predicted 2D landmarks are correct, our method can predict different canonical face shapes for different input views of the same subject as it only processes a single image at a time. The predicted landmarks and the canonical 3D shape for a frontal and profile image are shown in the first four columns. The change between the two predicted shapes is visualized as a heatmap in the last column (scale 0-10 mm).}
        \label{fig:changeInShape}
    \end{centering}
\end{figure}

We observe that our method can fail under strong head rotations as shown in the first row of \figref{fig:failureCases}. The spatial transformer can also have difficulties in localizing the face tightly when face occupies a small region in the input image (see second row in \figref{fig:failureCases}). Another limitation of our approach is that it is designed to handle only inputs containing a single subject, while a generic face detection algorithm can handle inputs with arbitrary number of faces. Finally as our network process a single image at a time, it can predict different canonical 3D shapes even when only the viewpoint of the input changes. In \figref{fig:changeInShape} we show how the predicted canonical shape changes for an input face in a profile view when compared to a frontal view. Restricting the predicted shape using a 3DMM or enforcing some multiview consistency during training might minimize these effects and produce more consistent geometry.

\section{Conclusion}
\label{sec:conclusion}

In this work we present Infinite 3D Landmarks, an improved method for continuous 2D facial landmark detection by introducing three architectural changes to a recent state-of-the-art landmark detector.  First, we add a spatial transformer network to automatically predict the facial bounding box, removing the need for offline face normalization.  Training this network alongside the landmark predictor optimizes the bounding box detection for our specific task.  Second, we modify the output head of the landmark predictor to estimate landmarks in a canonical 3D space, together with the head pose and camera focal length, allowing the network to reason about the 3D spatial layout of the landmarks and compute important metadata like landmark visibility.  Finally, we explicitly account for inconsistencies in landmark annotations across different training datasets by introducing a query deformer network, further improving the accuracy of the landmark prediction.   Our contribution is the combination of these three modifications, which we use to augment the baseline landmark detection method of Chandran~\etal~\cite{Chandran2023} and demonstrate significant improvements in accuracy and temporal stability.  Finally, the predicted 3D landmarks are also beneficial for downstream applications like 3D face reconstruction, texture completion and landmark visibility estimation.

\bibliographystyle{eg-alpha-doi}
\bibliography{infinite3dlandmarks}


\newpage

\end{document}